\documentclass[journal,onecolumn]{IEEEtran}
\ifCLASSINFOpdf
   \usepackage[pdftex]{graphicx}
\else
\fi
\usepackage{amsmath}
\usepackage{amssymb}
\usepackage[ruled,vlined]{algorithm2e}
\usepackage{hyperref}
\usepackage[justification=centering]{caption}
\begin{document}
%
\title{Weak Supervision with Incremental Source Accuracy Estimation}
%
%
%

\author{Richard~Correro,~\IEEEmembership{Department of Statistics,~Stanford University,}}

\maketitle

\begin{abstract}
Motivated by the desire to generate labels for real-time data we develop a method to estimate the dependency structure and accuracy of weak supervision sources incrementally. Our method first estimates the dependency structure associated with the supervision sources and then uses this to iteratively update the estimated source accuracies as new data is received. Using both off-the-shelf classification models trained using publicly-available datasets and heuristic functions as supervision sources we show that our method generates probabilistic labels with an accuracy matching that of existing off-line methods.
\end{abstract}


%
\IEEEpeerreviewmaketitle

\section{Introduction}
%
%
%
%
\IEEEPARstart{W}{eak} supervision approaches obtain labels for unlabeled training data using noiser or higher level sources than traditional supervision [1]. These sources may be heuristic functions, off-the-shelf models, knowledge-base-lookups, etc. [2]. By combining multiple supervision sources and modeling their dependency structure we may infer the true labels based on the outputs of the supervision sources. 
\subsection*{Problem Setup}
In the weak supervision setting we have access to a dataset $X = \{x_1, \dots, x_n\}$ associated with unobserved labels $Y = \{y_1, \dots, y_n\}, \ \ y_i \in \{1, \dots, k\}$ and a set of weak supervision sources $p_i(y|x), i = 1, \dots, m$. 

We denote the outputs of the supervision sources by $\lambda_1, \dots, \lambda_m$ and let $\mathbf{\lambda_j} = [\lambda_1 \ \lambda_2 \ \dots \ \lambda_m]^T$ denote the vector of labels associated with example $x_j$. The objective is to learn the joint density 
$$f(y, \mathbf{\lambda})$$
over the sources and the latent label. Using this we may estimate the conditional density

\begin{align*}
   f_{Y\mid\Lambda}(y|\mathbf\lambda) = \frac{f_{Y, \Lambda}(y, \mathbf\lambda)}{f_{\Lambda}(\lambda)}, \quad  f_{\Lambda}(\lambda) > 0. \tag{1}
\end{align*}

These sources may take many forms but we restrict ourselves to the case in which 
$\lambda_i \in \{0, \dots, k\}$ and thus the label functions generate labels belonging to the same domain as $Y$. Here $\lambda_i = 0$ indicates the $i^{th}$ source has not generated a label for this example. Such supervision sources may include heuristics such as knowledge base lookups, or pre-trained models.

\section{Related Work}
Varma et. al. [3] and Ratner, et. al. [4] model the joint distribution of $\lambda_1, \dots, \lambda_m, Y$ in the classification setting as a Markov Random Field
$$f_G(\lambda_1, \dots, \lambda_m, y) = \frac{1}{Z}\exp\left(\sum_{\lambda_i \in V}\theta_i \lambda_i + \sum_{(\lambda_i, \lambda_j) \in E}\theta_{i,j}\lambda_i \lambda_j + \theta_Y y + \sum_{\lambda_i \in V} \theta_{Y,y}y\lambda_i \right)$$
associated with graph $G=(V,E)$ where $\theta_{i,j} \ 1 \leq i,j \leq m+1$ denote the canonical parameters associated with the supervision sources and $Y$, and $Z$ is a partition function [here $V = \{\lambda_1, \dots, \lambda_m\} \cup \{Y\}$]. If $\lambda_i$ is not independent of $\lambda_j$ conditional on $Y$ and all sources $\lambda_k, \ k\in \{1, \dots, m\}\setminus\{i,j\}$, then $(\lambda_i, \lambda_j)$ is an edge in $E$.


Let $\Sigma$ denote the covariance matrix of the supervision sources and $Y$. To learn $G$ from the labels 
$$O = \{\lambda_i : \lambda_i = [\lambda_1, \dots, \lambda_m]^T; i = 1,\dots,n\}$$
and without the ground truth labels, Varma et. al. assume that $G$ is sparse and therefore that the inverse covariance matrix $\Sigma^{-1}$ associated with $\lambda_1, \dots, \lambda_m, Y$ is graph-structured. Since $Y$ is a latent variable the full covariance matrix $\Sigma$ is unobserved. We may write the covariance matrix in block-matrix form as follows:

$$Cov[O \cup S] := \Sigma = \begin{bmatrix} \Sigma_O & \Sigma_{OS} \\ \Sigma_{OS}^T & \Sigma_S \end{bmatrix}$$
Inverting $\Sigma$, we write
$$\Sigma^{-1} = \begin{bmatrix} K_O & K_{OS} \\ K_{OS}^T & K_S \end{bmatrix}$$
$\Sigma_O$ may be estimated empirically:
$$\hat\Sigma_O = \frac{\Lambda \Lambda^T}{n} - \nu\nu^T$$
where $\Lambda = [\mathbf{\lambda}_1 \mathbf{\lambda}_2, \dots, \mathbf{\lambda}_n]$ denotes the $m \times n$ matrix of labels generates by the sources and $\nu  = \hat{E}[O] \in \mathbb{R}^m$ denotes the observed labeling rates.

Using the block-matrix inversion formula, Varma et. al. show that
$$K_O = \Sigma_O^{-1} + c\Sigma_O^{-1}\Sigma_{OS}\Sigma_{OS}^T\Sigma_{O}^{-1}$$
where $c = (\Sigma_S - \Sigma_{OS}^T\Sigma_O^{-1}\Sigma_{OS})^{-1} \in \mathbb{R}^+$. Letting $z = \sqrt{c}\Sigma_{O}^{-1}\Sigma_{OS}$, they write
$$\Sigma_O^{-1} = K_O - z z^T$$
where $K_O$ is sparse and $z z^T$ is low-rank positive semi definite. Because $\Sigma_O^{-1}$ is the sum of a sparse matrix and a low-rank matrix we may use Robust Principal Components Analysis [5] to solve the following:
\begin{align*}
    & (\hat{S}, \hat{L}) = \text{argmin}_{(S,L)} ||L||_* + \gamma||S||_1 \\
    & s.t. \quad\quad S - L = \hat{\Sigma}^{-1}_O 
\end{align*}
Varma et. al. then show that we may learn the structure of $G$ from $K_O$ and we may learn the accuracies of the sources from $z$ using the following algorithm:

\begin{algorithm}[H]
\caption{Weak Supervision Structure Learning and Source Estimation Using Robust PCA (From [3])}
\SetAlgoLined
\KwResult{$\hat{G} = (V, \hat{E}), \ \hat{L}$}
$\mathbf{Inputs:}$ Estimate of covariance matrix $\hat\Sigma_O$, parameter $\gamma$, threshold $T$ \\
$\mathbf{Solve:} \quad (\hat{S}, \hat{L}) = \text{argmin}_{(S,L)} ||L||_* + \gamma||S||_1$ \\
s.t. $S - L = \hat{\Sigma}^{-1}_O$ \\
$\hat{E} \xleftarrow{}\{(i,j) : i < j, \hat{S}_{i,j} > T\}$
\end{algorithm}

Note that $\hat{L} = zz^T$.

Ratner, et. al. [4] show that we may estimate the source accuracies $\hat{\mu}$ from $z$ and they propose a simpler algorithm for estimating $z$ if the graph structure is already known: If $E$ is already known we may construct a dependency mask $\Omega = \{(i,j) : (\lambda_i, \lambda_j) \not\in E\}$. They use this in the following algorithm:

\begin{algorithm}[H]
\caption{Source Estimation for Weak Supervision (From [4])}
\SetAlgoLined
\KwResult{$\hat{\mu}$}
$\mathbf{Inputs:}$ Observed labeling rates $\hat{\mathbb{E}}[O]$ and covariance $\hat{\Sigma}_O$; class balance $\hat{\mathbb{E}}[Y]$ and variance $\hat{\Sigma}_S$; dependency mask $\Omega$\\
$\hat{z} \xleftarrow{} \text{argmin}_Z ||\hat{\Sigma}_{O}^{-1} + z z^T||_{\Omega}$ \\
$\hat c \xleftarrow{} \Sigma_{S}^{-1}(1 + \hat{z}^T \hat{\Sigma}_O \hat{z})$ \\
$\hat{\Sigma}_{OS} \xleftarrow{} \hat{\Sigma}_O \hat{z}/\sqrt{\hat{c}}$ \\
$\hat{\mu} \xleftarrow{} \hat{\Sigma}_{OS} + \hat{\mathbb{E}}[Y]\hat{\mathbb{E}}[O]$ \\
\end{algorithm}

Snorkel, an open-source Python package, provides an implementation of algorithm 2 [6].

\section{Motivating Our Approach}
Although the algorithm proposed by Varma et. al. may be used determine the source dependency structure and source accuracy, it requires a robust principal components decomposition of the matrix $\hat{\Sigma}_O$ which is equivalent to a convex Principal Components Pursuit (PCP) problem [5]. Using the current state-of-the-art solvers such problems have time complexity $O(\epsilon^{-2})$ where $\epsilon$ denotes the solver convergence tolerance [5]. For reasonable choices of $\epsilon$ this may be a very expensive calculation. 

In the single-task classification setting, algorithm 2 may be solved by least-squares and is therefore much less expensive to compute than algorithm 1. Both algorithms, however, require the observed labeling rates and covariance estimates of the supervision sources over the entire dataset and therefore cannot be used in an on-line setting.

We therefore develop an on-line approach which estimates the structure of $G$ using algorithm 1 on an initial "minibatch" of unlabeled examples and then iteratively updates the source accuracy estimate $\hat{\mu}$ using using a modified implementation of algorithm 2.

\section{Methods}
Given an initial batch $b_1$ of unlabeled examples $X_{b_1} = \{x_1, \dots, x_k\}$ we estimate $G$ by first soliciting  labels $\mathbf{\lambda}_1, \dots, \mathbf{\lambda}_k$ for $X_{b_1}$ from the sources. We then calculate estimated labeling rates $\hat{E}[O]$ and covariances $\hat{\Sigma}_{Ob_1}$ which we then input to algorithm 1, yielding $\hat{G} = (V, \hat{E})$ and $\hat{L}$. From $\hat{E}$ we create the dependency mask $\hat{\Omega} = \{(i,j) : (\lambda_1, \lambda_j) \not\in \hat{E}\}$ which we will use with future data batches. Using the fact that $\hat{L} = zz^T$ we recover $\hat{z}$ by first calculating
$$|\hat{z}| = \sqrt{diag(\hat{L})}$$
We then break the symmetry using the method in [4]. Note that if a source $\lambda_i$ is conditionally independent of the others then the sign of $z_i$ determines the sign of all other elements of $z$.

Using $\hat{z}, \ \hat{E}[O], \ \hat{\Sigma}_{Ob_1}$, class balance prior $\hat{E}[Y]$ and class variance prior $\hat{\Sigma}_S$ we calculate $\hat{\mu}$, an estimate of the source accuracies [if we have no prior beliefs about the class distribution then we simply substitute uninformative priors for $\hat{E}[O]$ and  $\hat{\Sigma}_{Ob_1}$].

For each following batch $b_p$ of unlabeled examples $X_{b_p}$ we estimate $\Sigma_{Ob_p}$ and $E[O]_{b_p}$. Using these along with $\hat{E}[O]$ and  $\hat{\Sigma}_{Ob_1}$ we calculate $\hat{\mu}_{b_p}$, an estimate of the source accuracies over the batch. We then update $\hat{\mu}$ using the following update rule:
$$\hat{\mu} := (1 - \alpha) \hat\mu + \alpha\mu_{b_p}$$
where $\alpha \in [0,1]$ denotes the mixing parameter. Our method thus models the source accuracies using an exponentially-weighted moving average of the estimated per-batch source accuracies.

Using the estimated source accuracies and dependency structure we may estimate $p(y, \mathbf{\lambda})$ which we may then use to estimate $p(y|\mathbf\lambda)$ by (1).

\begin{algorithm}[H]
\caption{Incremental Source Accuracy Estimation}
\SetAlgoLined
\KwResult{$\hat{\mu}$}
$\mathbf{Inputs:}$ Observed labeling rates $\hat{\mathbb{E}}[O]_b$ and covariance $\hat{\Sigma}_{Ob}$; class balance $\hat{\mathbb{E}}[Y]$ and variance $\hat{\Sigma}_S$ \\
\For{each batch b}{
\eIf{is initial batch}{
    Use algorithm 1 to calculate $\hat{G}$ and $\hat{L}$ \\
    $|\hat{z}| \xleftarrow{} \sqrt{diag(\hat{L})}$ \\
    Determine the sign of the entries of $z$ using method from [4] \\
}{
    $\hat{z} \xleftarrow{} \text{argmin}_z ||\hat{\Sigma}_{Ob}^{-1} + z z^T||_{\Omega}$ \\
}
$\hat c \xleftarrow{} \Sigma_{S}^{-1}(1 + \hat{z}^T \hat{\Sigma}_{Ob} \hat{z})$ \\
$\hat{\Sigma}_{OS} \xleftarrow{} \hat{\Sigma}_{Ob} \hat{z}/\sqrt{\hat{c}}$ \\
$\hat{\mu}_b \xleftarrow{} \hat{\Sigma}_{OS} + \hat{\mathbb{E}}[Y]\hat{\mathbb{E}}[O]_b$ \\
\eIf{is initial batch}{
    $\hat{\mu} \xleftarrow{} \hat{\mu}$ \\
}{
 $\hat\mu \xleftarrow{} (1 - \alpha)\hat\mu + \alpha \hat{\mu}_b$ \\   
}
}
\end{algorithm}

\section{Tests}
\subsection*{Supervision Sources}
We test our model in an on-line setting using three supervision sources. Two of the sources are off-the-shelf implementations of Naïve Bayes classifiers trained to classify text by sentiment. Each was trained using openly-available datasets. The first model was trained using a subset of the IMDB movie reviews dataset which consists of a corpus of texts labeled by perceived sentiment [either "positive" or "negative"]. Because the labels associated with this dataset are binary the classifier generates binary labels. 

The second classifier was trained using another openly-available dataset, this one consisting of a corpus of text extracted from tweets associated with air carriers in the United States and labeled according to sentiment. These labels in this dataset belong to three seperate classes ["positive", "neutral", and "negative"] and therefore the model trained using this dataset classifies examples according to these classes.

The final supervision source is the Textblob Pattern Analyzer. This is a heuristic function which classifies text by polarity and subjectivity using a lookup-table consisting of strings mapped to polarity/subjectivity estimates. To generate discrete labels for an example using this model we threshold the polarity/subjectivity estimates associated with the label as follows:
\begin{itemize}
    \item If polarity is greater than 0.33 we generate a positive label 
    \item If polarity is less than or equal to 0.33 but greater than -0.33 we generate a neutral label
    \item If polarity is less than or equal to 0.33 we generate a negative label
\end{itemize}

\subsection*{Test Data}
We test our incremental model using a set of temporally-ordered text data extracted from tweets associated with a 2016 GOP primary debate labeled by sentiment ["positive", "neutral", or "negative"]. We do so by solicting labels $\mathbf{\lambda}_1, \dots, \mathbf{\lambda}_n$ associated with the $n$ examples from the three supervision sources.

\subsection*{Weak Supervision as Transfer Learning}
Note that this setting is an example of a transfer learning problem [7]. Specifically, since we are using models pre-trained on datasets similar to the target dataset we may view the Naive Bayes models as transferring knowledge from those two domains [Tweets associated with airlines and movie reviews, respectively] to provide supervision signal in the target domain [7]. The Pattern Analyzer may be viewed through the same lens as it uses domain knowledge gained through input from subject-matter experts.


\subsection*{Test Setup}
Because our model is generative we cannot use a standard train-validation-test split of the dataset to determine model performance. Instead, we compare the labels generated by the model with the ground-truth labels over separate folds of the dataset.

\subsubsection*{Data Folding Procedure}
We split the text corpus into five folds. The examples are not shuffled to perserve temporal order within folds. Using these folds we perform 5 separate tests, each using four of the five folds in order. For example, the fifth test uses the fold 5 and folds 1---3, \textit{in that order}.

\subsubsection*{Partition Tests}
For each set of folds we further partition the data into $k = 100$ batches of size $q$ which we refer to as "minibatches" [as they are subsets of the folds]. For each minibatch we solicit labels $\mathbf{\lambda}_1, \dots, \mathbf{\lambda}_q, \ \mathbf{\lambda}_i \in \mathbf{R^3}$ from the two pretrained models and the Pattern Analyzer. Note that both pretrained classifiers first transform the text by tokenizing the strings and then calculating the term-frequency to inverse document frequency (Tf-idf) for each token. We store these labels in an array $\mathbf{L}$ for future use. We then calculate $\hat{E}[O]_b$ and  $\hat{\Sigma}_{Ob}$ for the minibatch, which we use with algorithm 3 to generate $\hat{\mu}_b$ and the dependency graph $\hat{G}$. Using these we generate labels corresponding to the examples contained within the minibatch.

Using the ground-truth labels associated with the examples contained within the minibatch we calculate the accuracy of our method by comparing the generated labels $\mathbf{\hat{y}}$ with the ground-truth labels $\mathbf{y}$:
$$
\texttt{accuracy}(\mathbf{y}, \mathbf{\hat{y}}) = \frac{1}{q} \sum_{i=0}^{q-1} \mathbf{1}(\mathbf{\hat{y}}_i = \mathbf{y}_i)
$$
We then average the accuracy scores associated with each minibatch over the number of minibatches used in each test to calculate the average per-test accuracy [calculated using four of the five folds of the overall dataset]. 

We then compare the average accuracies of the labels produced using our incremental method to the accuracies of the labels produced by an existing off-line source accuracy estimation method based on algorithm 2 [6]. Since this method works in an off-line manner it requires access to the entire set $\mathbf{L}$ of labels generated by the supervision sources. Using these this method generates its own set of generated labels $\mathbf{\hat{y}}_{baseline}$ with which we then calculate the baseline accuracy using the accuracy metric above.

Finally, we compare the accuracy of the labels generated by our method with the accuracy of the labels generated by each of the supervision sources.

\subsubsection*{Comparing Values of $\alpha$}
We then follow the same procedure as above to generate labels for our method, except this time we use different values of $\alpha$.

\section{Results}
Our tests demonstrate the following:
\begin{enumerate}
    \item Our model generates labels which are more accurate than those generated by the baseline [when averaged over all 5 tests].
    \item Both our method and the baseline generate labels which are more accurate than those generated by each of the supervision sources.
    \item Our tests of the accuracy of labels generated by our method using different values of $\alpha$ yields an optimal values $\alpha = 0.05$ and shows convexity over the values tested.
\end{enumerate}
Theses tests show that the average accuracy of the incremental model qualitatively appears to increase as the number of samples seen grows. This result is not surprising as we would expect our source accuracy estimate approaches the true accuracy $\hat{\mu} \xrightarrow{} {\mu}$ as the number of examples seen increases. This implies that the incremental approach we propose generates more accurate labels as a function of the number of examples seen, unlike the supervision sources which are pre-trained and therefore do not generate more accurate labels as the number of labeled examples grows.

These tests also suggest that an optimal value for $\alpha$ for this problem is approximately $0.05$ which is in the interior of the set of values tested for $\alpha$. Since we used $100$ minibatches in each test of the incremental model this implies that choosing an $\alpha$ which places greater weight on more recent examples yields better performance, although more tests are necessary to make any stronger claims.

Finally, we note that none of the models here tested are \textit{in themselves} highly-accurate as classification models. This is not unexpected as the supervision sources were intentionally chosen to be "off-the-shelf" models and no feature engineering was performed on the underlying text data, neither for the datasets used in pre-training the two classifier supervision sources nor for the test set [besides Tf-idf vectorization]. The intention in this test was to compare the relative accuracies of the two generative methods, not to design an accurate discriminative model.

\section{Conclusion}
We develop an incremental approach for estimating weak supervision source accuracies. We show that our method generates labels for unlabeled data which are more accurate than those generated by pre-existing non-incremental approaches. We frame our specific test case in which we use pre-trained models and heuristic functions as supervision sources as a transfer learning problem and we show that our method generates labels which are more accurate than those generated by the supervision sources themselves.

\begin{figure}[!t]
\captionsetup{justification=centering}
\centering
{\includegraphics[width=2.5in]{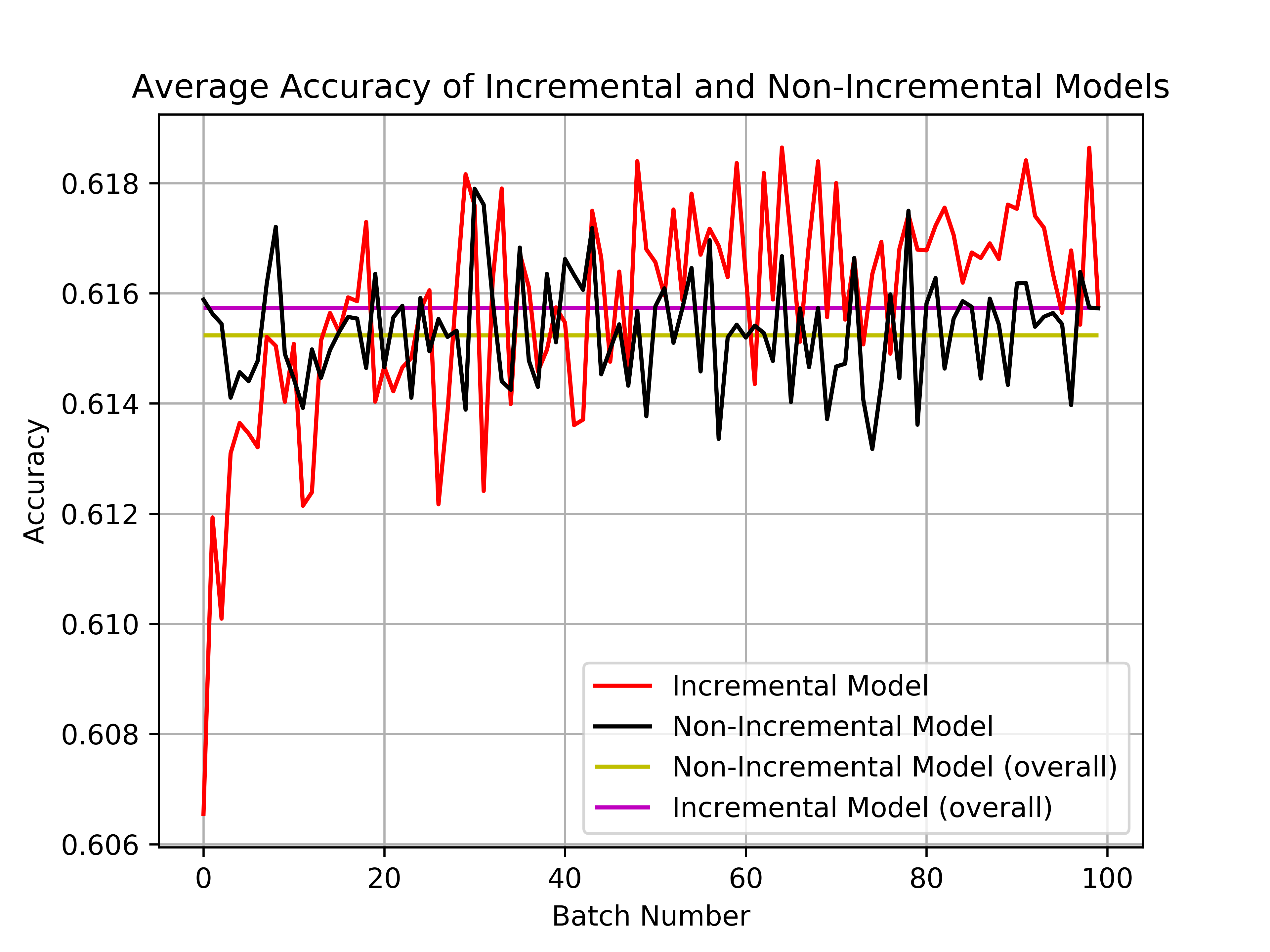}
\label{fig_first_case}}
\caption{Comparison of incremental and non-incremental model accuracy over minibatches.}
\label{fig_sim}
\end{figure}

\begin{figure}[!t]
\centering
{\includegraphics[width=2.5in]{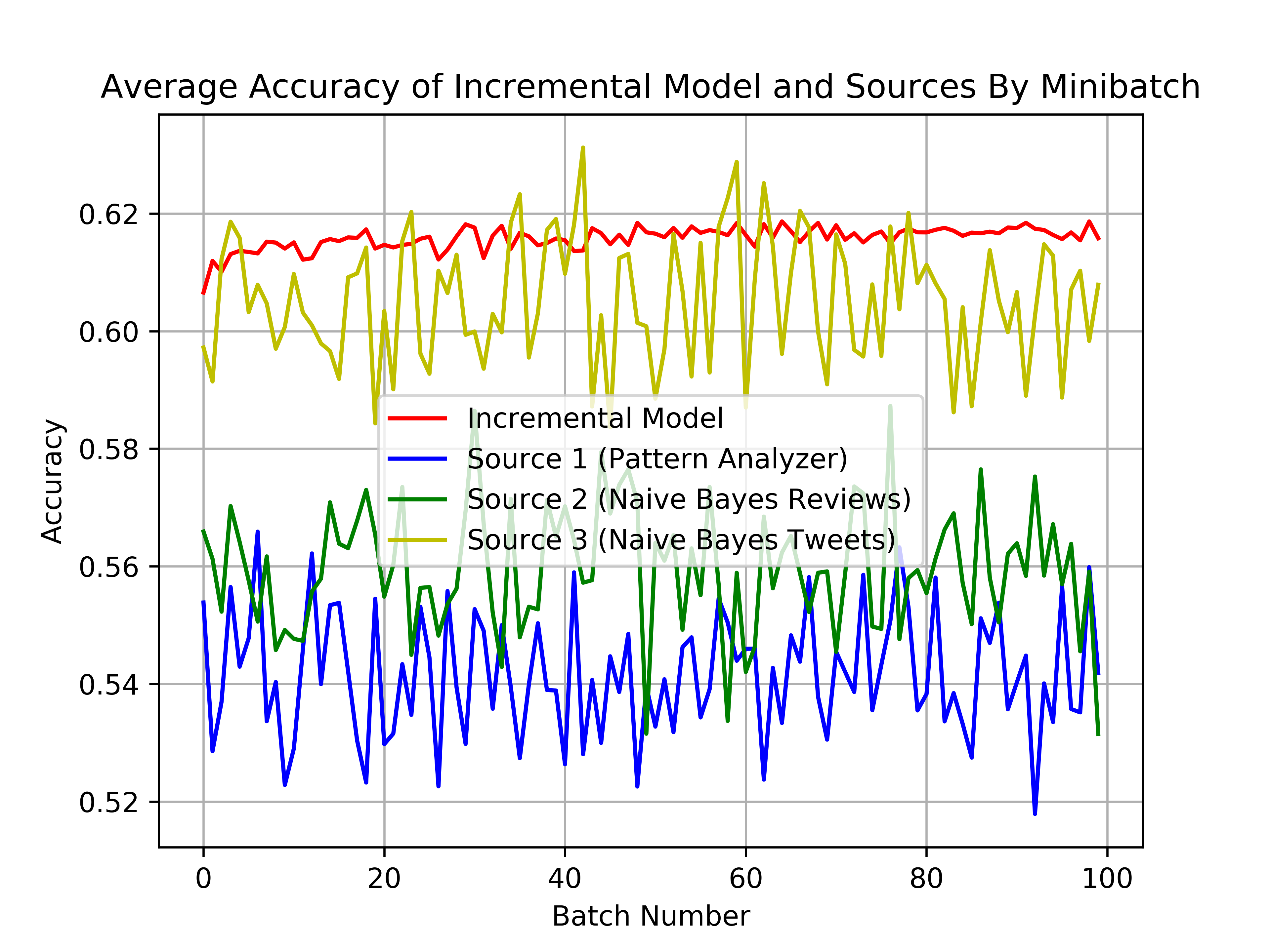}
\label{fig_second_case}}
\caption{Average model accuracy over minibatches.}
\label{fig_sim}
\end{figure}

%
%
\begin{figure}[!t]
\centering
\includegraphics[width=2.5in]{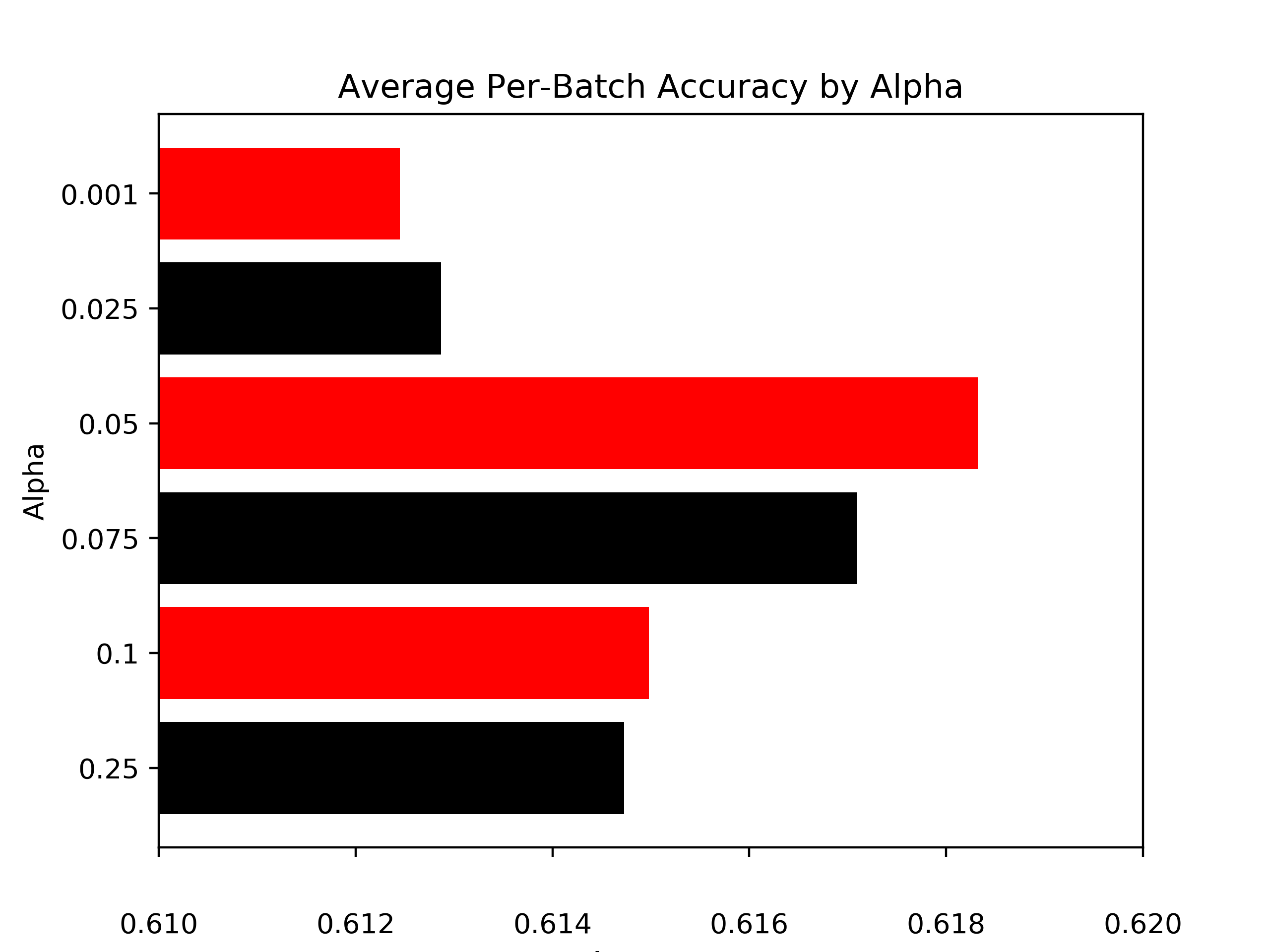}
\caption{Average per-batch accuracies for different values of $\alpha$.}
\label{fig_sim}
\end{figure}


%

%

%
\begin{table}[!t]
\renewcommand{\arraystretch}{1.3}
\caption{Average Accuracy of Incremental Model For Different Alpha Values}
\label{table_example}
\centering
\begin{tabular}{|c|c|c|c|c|c|c|}
\hline
Alpha & 0.001 & 0.01 & 0.025 & 0.05 & 0.1 & 0.25 \\
\hline
Accuracy & 0.61245 & 0.61287 & 0.61832 & 0.61709 & 0.61498 & 0.61473\\
\hline
\end{tabular}
\end{table}

\end{document}